\documentclass[letterpaper]{article} 
\usepackage{aaai23}  
\usepackage{times}  
\usepackage{helvet}  
\usepackage{courier}  
\usepackage[hyphens]{url}  
\usepackage{graphicx} 
\usepackage{natbib}  
\usepackage{caption} 
\usepackage{algorithm}
\usepackage{algorithmic}

\usepackage[export]{adjustbox}

\usepackage{newfloat}
\usepackage{listings}
\DeclareCaptionStyle{ruled}{labelfont=normalfont,labelsep=colon,strut=off} 
\floatstyle{ruled}
\newfloat{listing}{tb}{lst}{}
\floatname{listing}{Listing}
\title{StockEmotions: Discover Investor Emotions \\ for Financial Sentiment Analysis and Multivariate Time Series}
\author{
    Jean Lee, \textsuperscript{\rm 1}
    Hoyoul Luis Youn, \textsuperscript{\rm 3}
    Josiah Poon, \textsuperscript{\rm 1}
    Soyeon Caren Han \textsuperscript{\rm 1}\textsuperscript{\rm 2}\thanks{Corresponding Author (caren.han@sydney.edu.au)}
}
\affiliations {
    \textsuperscript{\rm 1} The University of Sydney,
    \textsuperscript{\rm 2} The University of Western Australia,
    \textsuperscript{\rm 3} KPMG Australia\\
    \{jean.lee, josiah.poon, caren.han\}@sydney.edu.au
    
}

\begin{document}

\maketitle

\begin{abstract}
There has been growing interest in applying NLP techniques in the financial domain, however, resources are extremely limited. This paper introduces StockEmotions, a new dataset for detecting emotions in the stock market that consists of 10k English comments collected from StockTwits, a financial social media platform. Inspired by behavioral finance, it proposes 12 fine-grained emotion classes that span the roller coaster of investor emotion. Unlike existing financial sentiment datasets, StockEmotions presents granular features such as investor sentiment classes, fine-grained emotions, emojis, and time series data. To demonstrate the usability of the dataset, we perform a dataset analysis and conduct experimental downstream tasks. For financial sentiment/emotion classification tasks, DistilBERT outperforms other baselines, and for multivariate time series forecasting, a Temporal Attention LSTM model combining price index, text, and emotion features achieves the best performance than using a single feature. 
\end{abstract}

\section{Introduction}

Natural Language Processing (NLP) in finance is a growing research area and has attracted considerable attention from both academics and industry. Researchers are seeking to analyze financial sentiment collected from social media \cite{cortis2017semeval, chen2020issues, xing2020financial} or news \cite{du2020stock, lee2021fednlp}, and combine financial text mining with historical price data for stock market prediction \cite{sawhney2020deep}. In particular, public mood and financial sentiment play a significant role in investment decisions as is explored in behavioral finance \cite{griffith2020emotions, zaleskiewicz2020emotions}; furthermore, social media messages have been studied as useful resources for detecting sentiment in NLP research \cite{ge2020beyond}.
 
However, existing datasets are (1) very limited in availability, mostly small, and even containing empty labels in the training set \cite{maia201818}. In addition, there is (2) no text data available that include investors’ emotions for stock market time series prediction. For example, existing studies use a proxy of public mood to predict the stock market \cite{bollen2011twitter} instead of extracting information from text data or applying an emotion taxonomy. 

\begin{figure*}[t]
  \centering
  \includegraphics[width=.9\linewidth]{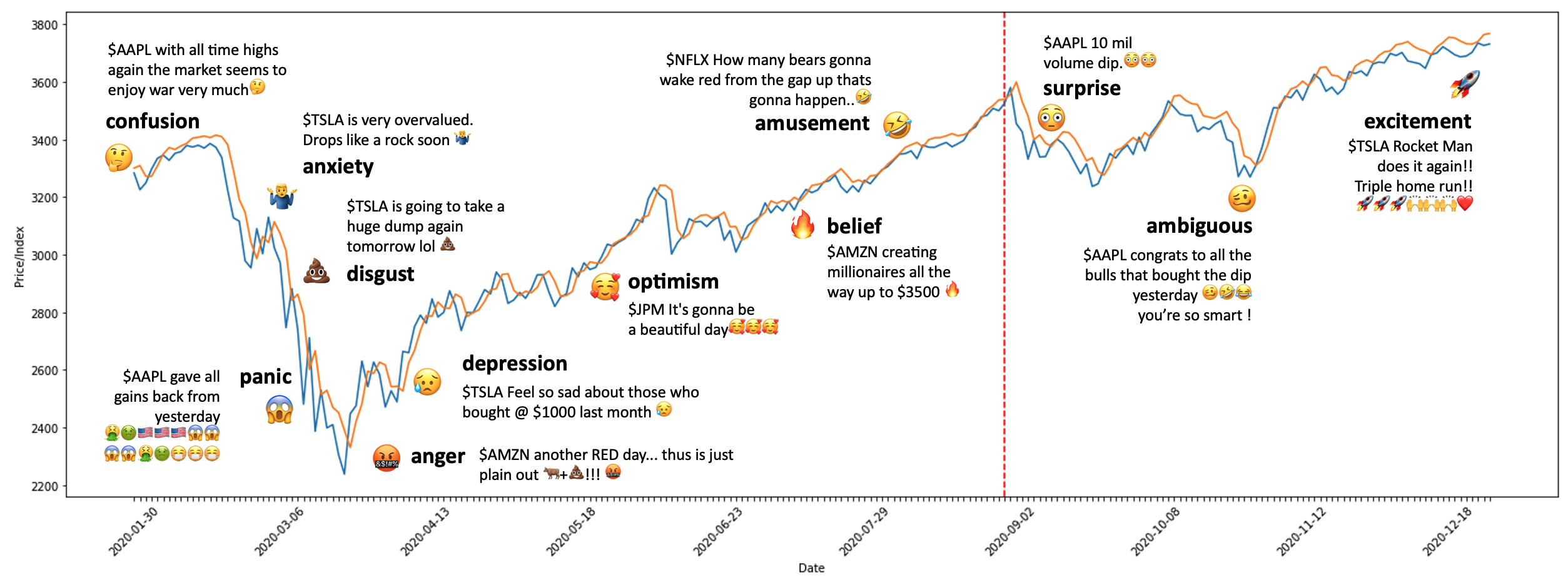}
  \caption{Example from StockEmotions dataset showing investor psychology on the stock market. A combination of input data (stock price index, text and emoji, and emotion label) is used on a Temporal Attention LSTM for multivariate time series forecasting. {\itshape (blue line = the actual S\&P index, orange line = the prediction S\&P index with a rolling window size 5, red vertical line = the point of data splitting; Further details are in the experiments section.)}}
  \label{fig:timeseries}
\end{figure*}

Inspired by behavioral finance \cite{hens2015behavioral}, we introduce StockEmotions, a dataset for emotion classification in the stock market, composed of 10,000 sentences collected from StockTwits. Our dataset contains 2 financial sentiment classes annotated by users who share the comments as well as 12 emotion classes by leveraging a pre-trained language model (PLM) and finance experts. We design an emotion taxonomy associated with existing psychology studies \cite{zaleskiewicz2020emotions, li2021past} to maximize the impact of the dataset in the financial domain. Figure \ref{fig:timeseries} shows samples from our dataset. For example, the comment “\$TSLA rocket man, does it again! Triple home run!\includegraphics[scale=0.06]{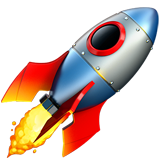}\includegraphics[scale=0.06]{Emoji/rocket_1f680.png}\includegraphics[scale=0.06]{Emoji/rocket_1f680.png}” has the {\itshape bullish} financial sentiment class, the {\itshape excitement} emotion class, and time series data and emoji data. Unlike the existing datasets \cite{cortis2017semeval, maia201818, xu2018stock}, StockEmotions provides diversified emotions both positive and negative sentiment.

In order to demonstrate the usability of the dataset, we conduct a dataset analysis and present baseline models for downstream tasks. For financial sentiment analysis, we implement seven baseline models including GRU \cite{cho2014learning}, DistilBERT \cite{sanh2019distilbert}, BERT \cite{DevlinCLT19}, and RoBERTa \cite{liu2019roberta}. DistilBERT outperforms other baselines, achieving an average F1-score of 0.81 for financial sentiment classification and 0.42 for emotion classification, respectively. For time series forecasting, we implement a modified Temporal Attention LSTM \cite{windsor2022improving} in order to identify the importance of text and emotion for the prediction. When the model jointly learns from stock price/index, text, and emotion features, we achieve the best performance on S\&P 500 compared to having numeric input data only. The major contributions of this study are as follows: 

\begin{itemize}
\setlength\itemsep{0em}
\item We create StockEmotions, a financial-domain-focused dataset for financial sentiment/emotion classification and stock market time series prediction; 
\item We apply a multi-step annotation pipeline that brings the collaboration of human and pre-trained language model; 
\item We demonstrate the dataset usability through downstream tasks and the impact of investor emotions for time series forecasting in particular.
\end{itemize}

\section{Related Work}

Emotion classification has been widely studied in NLP research \cite{buechel2021towards, park2021dimensional, hu2021bidirectional}, with data collected from social networks \cite{abdul2017emonet, mohammad2018semeval}, online news \cite{lei2014towards}, or dialog \cite{hsu2018emotionlines, chatterjee2019semeval}. Most existing studies contain annotations from two main emotion categories proposed by Ekman’s 6 emotions \cite{ekman1992argument} and Plutchik’s 8 emotions \cite{plutchik1980general}. A recent study proposes the largest human-annotated emotion dataset, GoEmotions \cite{demszky2020goemotions}, labeled for 27 emotion categories or neutral. In this work, BERT outperforms other baselines and further research shows similar performance \cite{alvarez2021uncovering}. Also, several existing datasets from social media focus on informal languages such as slang, emoji, and hashtags \cite{felbo2017using, shoeb2020emotag1200}. 

In the financial domain, analyzing the investor’s emotion has been studied \cite{duxbury2020emotions}; however, the availability of datasets is very limited. Existing datasets collected from StockTwits \cite{oliveira2016stock, li2017learning} have relatively small size of data and only focus on financial sentiment instead of considering in-depth emotion features in the semantic space. In addition, most studies for the stock market prediction using twitter applied an investor mood index \cite{bollen2011twitter, ge2020beyond} instead of extracting investor emotions from the collected text, or disregarding emotions \cite{xu2018stock}.

\section{StockEmotions Dataset}
\subsection{Data Collection}
StockTwits is a social media platform similar to Twitter but focused on the stock market. The users share short comments about companies with special features such as a cashtag (e.g. \$TSLA) and financial sentiment (e.g. bullish and bearish) annotated by the users. The collected raw data contains over 3 million comments that cover over 80\% of the S\&P 500 by market-capitalization-weighted. The date range is between 01 January 2020 and 31 December 2020 which covers a roller coaster of investor emotion in the era of COVID-19 \cite{naseem2021investor}.


\subsection{Data Processing}
In order to have fair representation and reduce user biases, we curate the comments shared by one user per day. To remove advertising or compromised data, we identify commercial usernames and commercial content text patterns and remove them using our best judgment.

\paragraph{\textbf{Tokenization and Length Filtering}}
We limit the sequence length to a maximum of 512 tokens to use BERT \cite{DevlinCLT19}. Also, we choose comments over 3 tokens long using NLTK’s word tokenizer and normalize the token repetitions by having over 4 unique tokens long. This length filtering is applied because short comments normally less convey contextual information.

\paragraph{\textbf{Masking}}
We use special characters for masking, such as a cashtag with a [CTAG] token, a hashtag with a [HTAG], or a website URL with a [URL]. When a comment contains only masked tokens, it is also removed.

\paragraph{\textbf{Handling Emoji}}
Users often express their emotions using emojis in social media and therefore we narrow down the comments consisting of at least one emoji. For instance, in this comment {\itshape "\$AAPL, \includegraphics[scale=0.06]{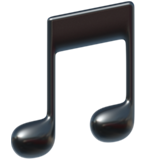} this is how we do it \includegraphics[scale=0.06]{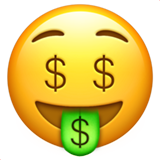} \includegraphics[scale=0.06]{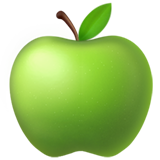}"}, it is difficult to detect investor’s emotion with text only. Thus, we convert emoji to its textual meaning. 



\subsection{Annotation} \label{annotation}

\begin{figure}[t]
  \centering
  \includegraphics[width=\linewidth]{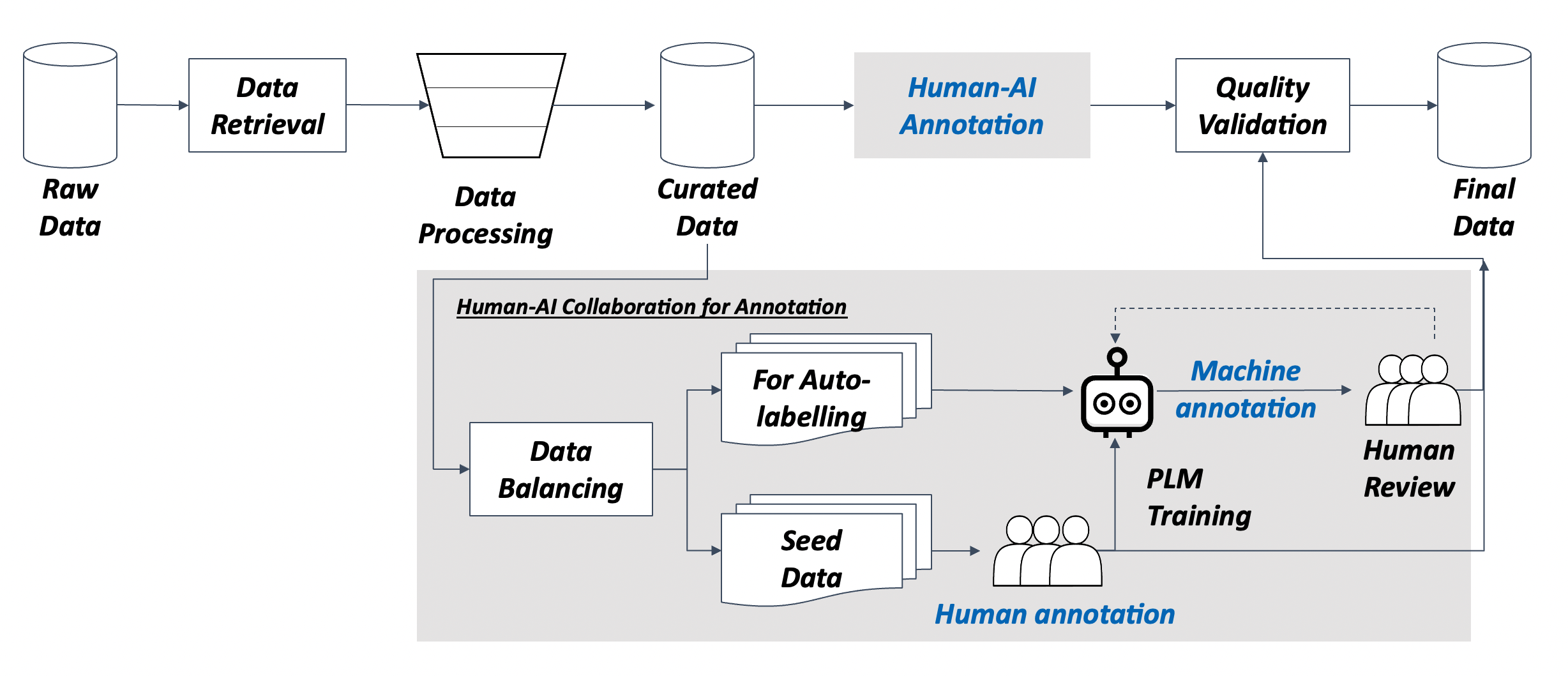}
  \caption{An overview of dataset creation pipeline. }
  \label{fig:data_flow}
\end{figure}

To annotate our data, we apply a multi-step pipeline inspired by human and machine collaboration \cite{nie2020adversarial}, which brings the combined capabilities of a pre-trained language model (PLM) and human annotators. As shown in Figure \ref{fig:data_flow}, we sample the seed dataset which has balanced sentiment classes and recruit three financial experts to annotate it. The taxonomy of emotions is updated based on feedback from annotators and empirical studies \cite{hens2015behavioral}. In the seed dataset (30\% of the final dataset), we achieve an average Cohen Kappa score of 0.79 \cite{cohen1960coefficient}. We construct a multi-classification language model by fine-tuning BERT on the automatic labeling data that is proportionally selected from each month. For the final dataset, annotators are asked to choose either revise or agree on the automatically labeled emotion. We update the fine-tuning BERT for the revised label and iterate the process.



\paragraph{\textbf{Data Quality Validation}}
We use a human-in-the-loop validation process to check the data quality compared to the GoEmotions \cite{demszky2020goemotions} whether the existing method can detect emotions in the financial content. When using the GoEmotions APIs on StockEmotions dataset, over 50\% of content is labeled as neutral, and the following labels are amusement (8.36\%), admiration (5.26\%), and joy (4.40\%). Several comments labeled as neutral using the GoEmotions are provided to three human evaluators who have not participated in the annotation process. All agreed that the comments contain investors' emotions which shows that the existing methods are not enough to detect emotions in the financial context, whereas our methods do.

\section{Data Analysis}

Table\ref{tab:stats} shows key statistics for the dataset. Our dataset provides 2 financial sentiment classes, 12 emotions, and time series data for the 10,000 comments. For the sentiment classes, we make a balanced dataset that consists of 55\% of bullish and 45\% of bearish. For the emotion classes, the distribution is imbalanced: {\itshape optimism, excitement, anxiety, and disgust} appear most frequently, whereas {\itshape panic, surprise, and depression} appear rarely. We also find that the users often share their wishes, hopes, or achievements even in the market downturn.

\begin{table}
\centering
\small
\setlength{\tabcolsep}{1mm}{
\renewcommand{\arraystretch}{1.2}
\begin{tabular}{lll}
\hline
\textbf{Number of Utterance} & 10,000  \\ \hline
\textbf{Number of Sentiment} & 2 - bullish (55\%), bearish (45\%) \\ \hline
\textbf{Number of Emotion}
& \begin{tabular}[c]{@{}l@{}}12 - ambiguous(9\%), amusement(8\%), \\
anger(4\%), anxiety(14\%), belief(9\%), \\ 
confusion(6\%), depression(2\%), \\ 
disgust(13\%), excitement(14\%), \\
optimism (16\%),  panic(3\%),  surprise(3\%) \end{tabular} \\ \hline
\textbf{Avg. Length} & 19.2 tokens per utterance \\ 
\textbf{Unique Emoji} & 761  \\ \hline
\textbf{Time Period} & 01 Jan 2020 - 31 Dec 2020 \\ \hline
\end{tabular}
}
\caption{Key statistics of StockEmotions. Each label shows the proportion in the total dataset.}
\label{tab:stats}
\end{table}

\begin{figure}[t]
  \centering
  \includegraphics[width=.75\linewidth]{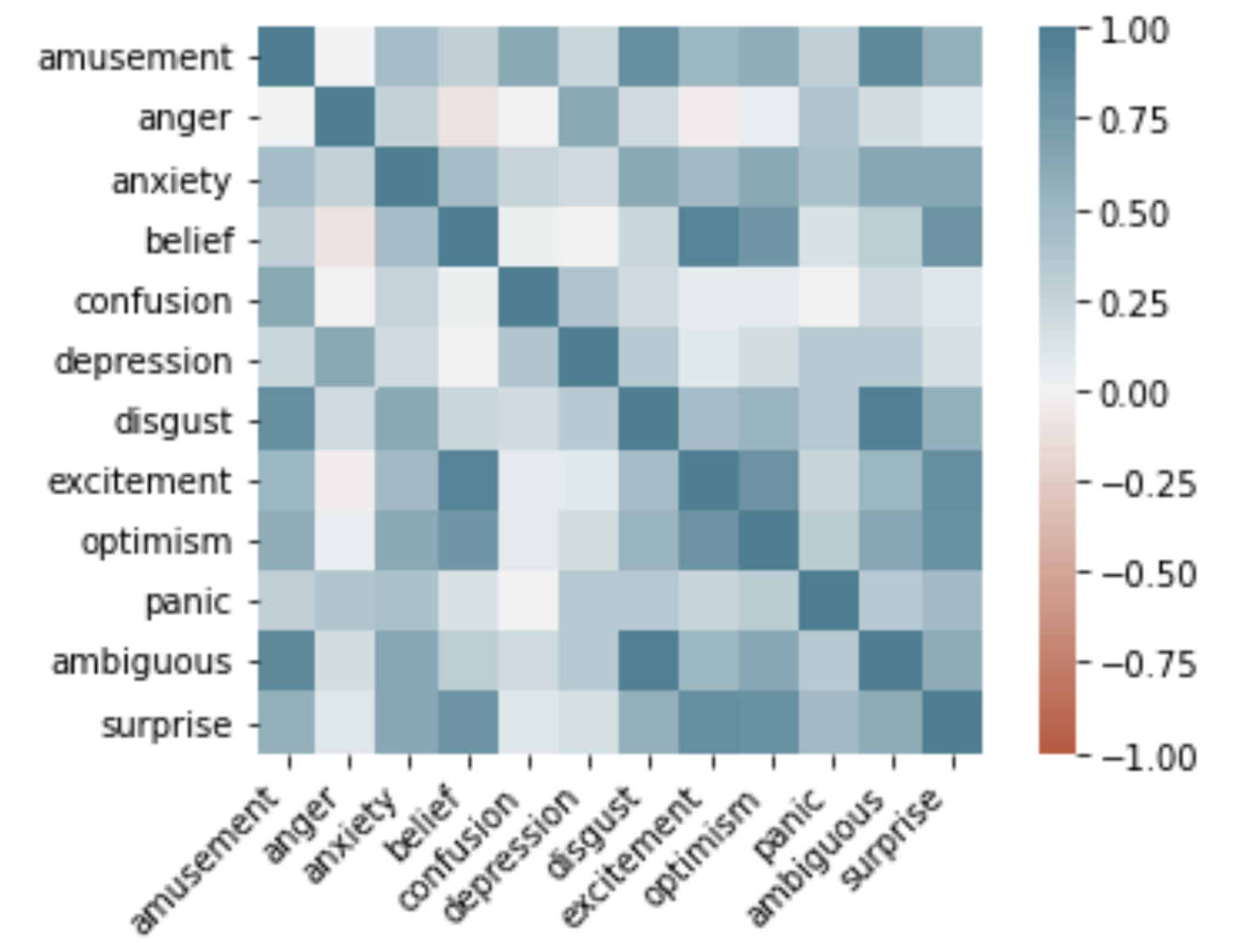}
  \caption{Emotions Correlation}
  \label{fig:emotion_corr}
\end{figure}

\paragraph{\textbf{Emotion Correlation}}
We study which emotions are likely to appear in the same context by examining their correlations. The heatmap in Figure \ref{fig:emotion_corr} shows a strong positive correlation in the emotion pairs, including {\itshape optimism-excitement} and {\itshape ambiguous-disgust}, while {\itshape anger-belief} are negatively correlated. We notice that teasing is very common on StockTwits, where users joke about others’ investing losses. Depending on the target of teasing, it can be labeled as {\itshape ambiguous} or {\itshape disgust}.

\paragraph{Emoji Analysis}
We further investigate how emojis are distributed along with sentiments and emotions. Interestingly, six emojis appear in the top 10 emoji rankings in both positive and negative sentiment. For example, the emoji \includegraphics[scale=0.06]{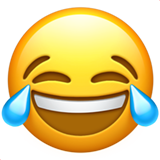} and \includegraphics[scale=0.06]{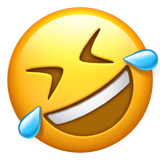} are used to not only express {\itshape excitement} but also convey laughing in {\itshape disgust}. 



\section{Modelling and Experiments}

\subsection{Financial Sentiment Analysis}
For the classification task, we randomly split the data into train/validation/test sets in the proportions of 80\%/10\%/10\%. For baseline experiments, we compare machine learning methods; (1) \textbf{Logistic Regression}, (2) \textbf{Naïve Bayes SVM} \cite{wang2012baselines}; standard neural network (3) \textbf{GRU} \cite{cho2014learning}, (4) \textbf{Bi-GRU}; and pre-trained language models; (6) \textbf{DistilBERT} \cite{sanh2019distilbert}, (7) \textbf{BERT}\(_{base}\) \cite{DevlinCLT19}, and (8) \textbf{RoBERTa}\(_{base}\) \cite{liu2019roberta}. 



\begin{table*}[t]
\centering
\small
\setlength{\tabcolsep}{1.2mm}{
\renewcommand{\arraystretch}{1.2}
\begin{tabular}{l|rrr|rrrrrrrrrrrrr}
\hline
\textbf{Model} & \multicolumn{3}{c|}{\textbf{Sentiment}} & \multicolumn{13}{c}{\textbf{Emotion}} \\ 
\cline{2-17}
\textit{(F1-score)}& bear. & bull. & \multicolumn{1}{|l|}{\textbf{avg.}} 
& ambg. & amus. & angr. & anxt. & belf. &  cnfs. & dprs. & disg. & exct. & optm. & panc. & surp. & \multicolumn{1}{|l}{\textbf{avg.}} \\ 
\hline
\textbf{LogitReg}. & 0.71 & \multicolumn{1}{r|}{0.77} & \textbf{0.74} 
& 0.12 & 0.29  & 0.44  & 0.37  & 0.29  & 0.48
& 0.24 & 0.29  & 0.39  & 0.31  & 0.24  & \multicolumn{1}{r|}{0.15}  & \textbf{0.32} \\ 
\textbf{NBSVM}. & 0.71 & \multicolumn{1}{r|}{0.78} & \textbf{0.75} 
& 0.10 & 0.27  & 0.45  & 0.30  & 0.36  & 0.46  
& 0.21 & 0.34  & 0.42  & 0.29  & 0.29  & \multicolumn{1}{r|}{0.22}  & \textbf{0.33} \\ 
\hline
\textbf{GRU}. & 0.72 & \multicolumn{1}{r|}{0.79} & \textbf{0.76} 
& 0.20 & 0.31  & 0.21  & 0.41  & 0.15  & 0.46  
& 0.19 & 0.33  & 0.39  & 0.38  & 0.43  & \multicolumn{1}{r|}{0.06}  & \textbf{0.34} \\ 
\textbf{Bi-GRU}. & 0.73 & \multicolumn{1}{r|}{0.78} & \textbf{0.76} 
& 0.22 & 0.33  & 0.49 & 0.39  & 0.30  & \textbf{0.54}  
& \textbf{0.29} & 0.39  & 0.43  & 0.32  & 0.41  & \multicolumn{1}{r|}{0.06}  & \textbf{0.36} \\ 
\hline
\textbf{DistilBERT}. & \textbf{0.79} & \multicolumn{1}{r|}{\textbf{0.83}} & \underline{\textbf{0.81}} 
& 0.12 & \textbf{0.37}  & 0.56  & 0.42  & \textbf{0.42}  & 0.51  
& \textbf{0.29} & \textbf{0.43}  & \textbf{0.51}  & \textbf{0.42}  & \textbf{0.48}  & \multicolumn{1}{r|}{0.21}  & \underline{\textbf{0.42}} \\ 
\textbf{BERT}. & \textbf{0.79} & \multicolumn{1}{r|}{\textbf{0.83}} & \underline{\textbf{0.81}} 
& \textbf{0.27} & 0.30  & \textbf{0.59} & \textbf{\textbf{0.46}}  & 0.37  & 0.50  
& 0.22 & 0.37  & 0.48  & 0.40  & 0.41  & \multicolumn{1}{r|}{0.41}  & \textbf{0.40} \\ 
\textbf{RoBERTa}. & 0.78 & \multicolumn{1}{r|}{0.82} & \textbf{0.80} 
& 0.09 & 0.25  & 0.13 & 0.44  & 0.29  & 0.50  
& 0.21 & \textbf{0.43}  & 0.44  & 0.39  & 0.11  & \multicolumn{1}{r|}{0.21}  & \textbf{0.39} \\ 
\hline
\end{tabular}
}
\caption{F1-score\(_{micro}\) results on the test set for classification task. Emotion labels are in alphabetical order; {\itshape ambiguous(ambg.), amusement(amus.), anger(angr.), anxiety(anxt.), belief(belf.), confusion(cnfs.), depression(dprs.), disgust(disg.), excitement(exct.), optimism(optm.), panic(panc.), and surprise(surp.)}.}
\label{tab:eval}
\end{table*}

\paragraph{Results of Financial Sentiment Analysis}
Table \ref{tab:eval} shows classification results in terms of the F1-score obtained from baseline models. For \textbf{financial sentiment (binary) classification}, the results provide a robust baseline, achieving  an average F1 score from 0.74 to 0.81 across the models. For \textbf{emotion (multi-class) classification}, DistilBERT achieved the best performance results (F1-score = 0.42) across the full taxonomy, similar to the BERT performance on GoEmotions (F1-score = 0.46) \cite{demszky2020goemotions}. Due to the imbalanced data, we found that less frequent emotions, such as \textit{ambiguous} and \textit{surprise} are likely to be confused. 


\subsection{Multivariate Time Series}
\begin{figure}[t]
  \centering
  \includegraphics[width=\linewidth]{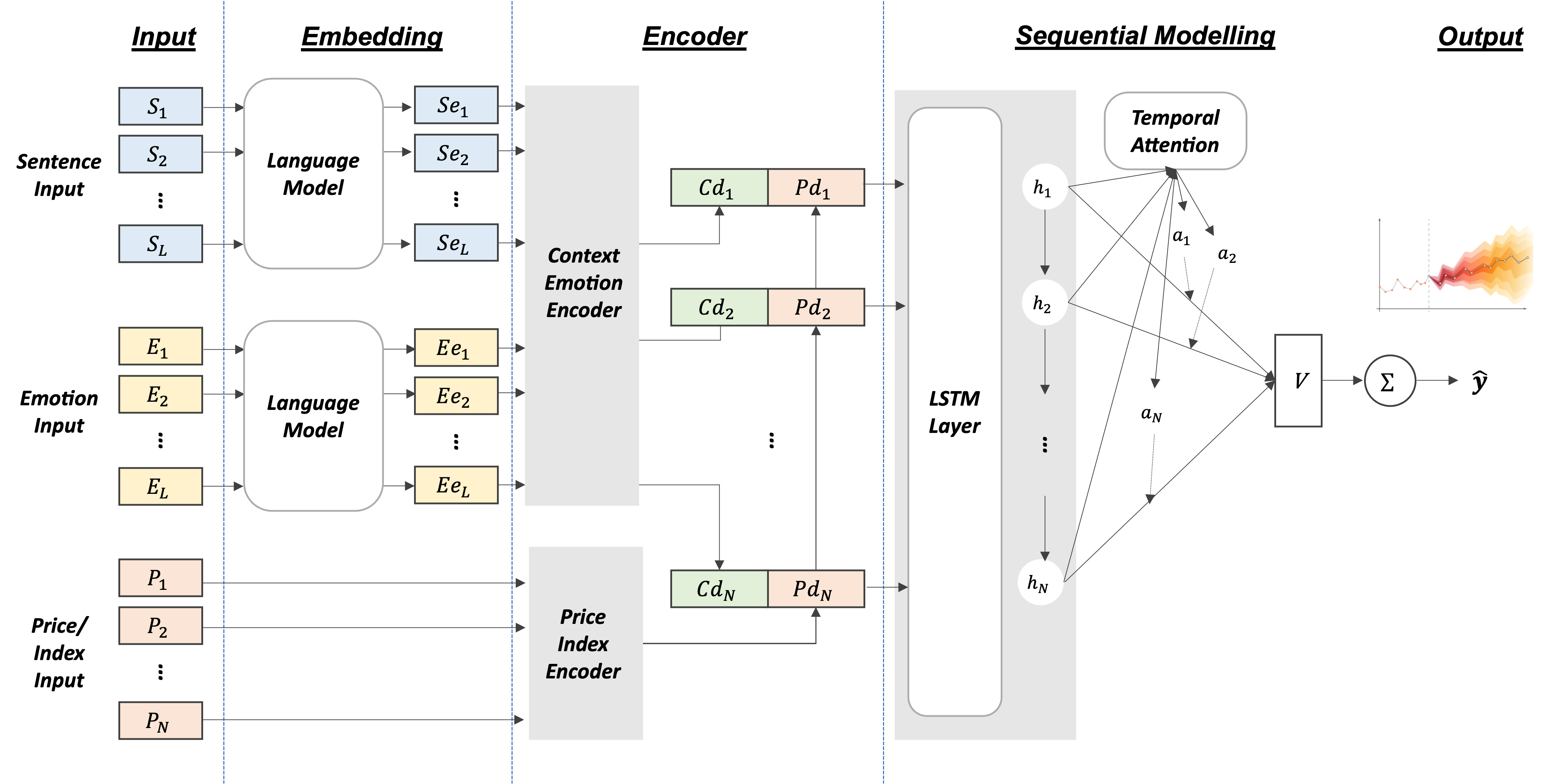}
  \caption{An overview of Temporal Attention LSTM.}
  \label{fig:model}
\end{figure}

\paragraph{Methodology}
Inspired by deep learning models in time series \cite{hu2018listening, hu2020deep, windsor2022improving}, we implement a Temporal Attention LSTM with bidirectional encoder representations from transformers (BERT). As shown in Figure \ref{fig:model}, the model learns temporally relevant information from sentences, emotions, and numerical data. To predict the price index of the S\&P 500, the model incorporates the numerical modality from the price index and the contextual modality from sentence embedding using BERT as well as emotion embedding using GloVe \cite{pennington2014glove} in a unified sequence. All embedding is encoded based on the intra-day sequence, returned as the context vector, concatenated with processed price index data, and fed into the Temporal Attention LSTM.

\begin{table}[t]
\centering
\small
\setlength{\tabcolsep}{1.8mm}{
\renewcommand{\arraystretch}{1.3}
\begin{tabular}{l|rrr|rrr}
\hline
\textbf{Model}
& \multicolumn{3}{c|}{\textbf{window size = 3}} 
& \multicolumn{3}{c}{\textbf{window size = 5}} \\ 
\cline{2-7}
hidden size
& \multicolumn{1}{c}{25} 
& \multicolumn{1}{c}{50} 
& \multicolumn{1}{c|}{100} 
& \multicolumn{1}{c}{25} 
& \multicolumn{1}{c}{50} 
& \multicolumn{1}{c}{100} \\ 
\hline
\textbf{only index}     
& \multicolumn{1}{r}{1.13} & 1.53 & 1.83
& \multicolumn{1}{|r}{1.15} & 2.07 & 1.99 \\ 
\textbf{ + text}       
& \multicolumn{1}{r}{2.18} & 2.30 & 1.49
& \multicolumn{1}{|r}{0.89} & 1.32 & 1.53 \\ 
\textbf{ + text + emo.} 
& \multicolumn{1}{r}{\textbf{1.06}} & \textbf{1.00} & \textbf{1.39}
& \multicolumn{1}{|r}{\underline{\textbf{0.83}}} & \textbf{1.08} & \textbf{1.48} \\ 
\hline
\end{tabular}
}
\caption{Mean Squared Error (MSE $*10^{-3}$) results on S\&P 500 and StockEmotions using a Temporal Attention LSTM model for time series \textit{(epochs = 250, emo. = emotion features)}.}
\label{tab:pred_error}
\end{table}

\paragraph{Experimental Setup}
For stock price/index data, we use a Yahoo finance APIs and split the data in the period of 01 Jan. 2020 - 03 Sep. 2020 for the training set (the former 67\% of total) and the data in the period of 04 Sep. 2020 - 31 Dec. 2020 for the test set (the remaining 33\% of total). For the performance evaluation, we use the Mean Squared Error (MSE; Lower values are better.) which measures the averaged squared difference between the predicted price and the actual value. Our hyper-parameter settings are as follows: a rolling window size [2, 3, \textbf{5}, 10, 15, 20, 25, 30]; hidden size [\textbf{25}, 50, 100]; and epochs [100, 150, 200, \textbf{250}] (The best performance is highlighted in bold.). A window size of 5 means that we predict the stock price of the next trading day based on combined historical data from the previous 5 days. 


 



\paragraph{Results of Multivariate Time Series}
Table \ref{tab:pred_error} shows the S\&P 500 index time series prediction in terms of MSE results across the different inputs. We found that the best performing results (MSE = 0.83*$10^{-3}$) are on S\&P 500 when combining price index, text, and emotion features with a rolling window size of 5. The experiment shows that adding text features to the index improves the prediction compared with using only numeric features. Moreover, adding emotion features to text and numeric features achieves the best performance, indicating the effectiveness of emotions in stock market forecasting.

\section{Conclusion}
We present StockEmotions, a financial-domain-focused dataset for financial sentiment classification and time series forecasting. Our dataset analysis and experiments on several downstream tasks show the usability of the dataset. We present the preliminary architecture on multivariate times series in this research. Further details and other baselines remain for future research.

\bibliography{aaai23}

\section{Acknowledgments}
This research is supported by an Australian Government Research Training Program (RTP) Scholarship.
The dataset can be found at \url{https://github.com/adlnlp/StockEmotions}.


\appendix
\section*{Appendix} \label{appendix}

\begin{table*}[ht]
\centering
\small
\setlength{\tabcolsep}{1.5mm}{
\renewcommand{\arraystretch}{1.2}
\begin{tabular}{lllll}
\hline
\textbf{Emotion}     & \textbf{Definition}                                                                                                                                                    & \textbf{Synonyms}                                                                                          & \textbf{Emoji} \\ 
\hline
\textbf{amusement}   & \begin{tabular}[c]{@{}l@{}}the pleasure that you get from being entertained or \\ from doing something interesting.\end{tabular}                                       & \begin{tabular}[c]{@{}l@{}}enjoyment, delight, \\ laughter, pleasure, fun\end{tabular}            & \includegraphics[scale=0.06]{Emoji/face-with-tears-of-joy_1f602.png}               \\ \hline
\textbf{anger}       & \begin{tabular}[c]{@{}l@{}}a strong feeling of being upset or annoyed \\ because of something wrong, unfair, cruel, or unacceptable.\end{tabular}                      & \begin{tabular}[c]{@{}l@{}}rage, outrage, fury,\\ wrath, irritation\end{tabular}                  & \includegraphics[scale=0.06]{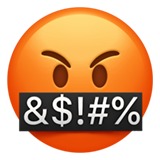}               \\ \hline
\textbf{anxiety}     & a feeling of nervousness or worry about what might happen                                                                                                              & \begin{tabular}[c]{@{}l@{}}nervousness, alarm, worry,\\ tension, uneasiness\end{tabular}          & \includegraphics[scale=0.06]{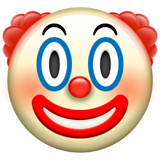}              \\ \hline
\textbf{belief}      & \begin{tabular}[c]{@{}l@{}}a feeling of certainty that something exists, is true, or is good,\\ associated with the company's operation.\end{tabular}                      & \begin{tabular}[c]{@{}l@{}}trust, faith, confidence,\\ conviction, reliance\end{tabular}          & \includegraphics[scale=0.06]{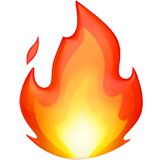}              \\ \hline
\textbf{confusion}   & a refusal or reluctance to believe                                                                                                                                     & \begin{tabular}[c]{@{}l@{}}scepticism, doubt, disbelief,\\ distrust, uncertainty\end{tabular}     & \includegraphics[scale=0.06]{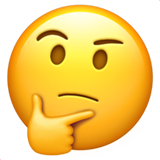}               \\ \hline
\textbf{depression}  & \begin{tabular}[c]{@{}l@{}}a state of feeling sad, extreme gloom, inadequacy, \\ and inability to concentrate\end{tabular}                                             & \begin{tabular}[c]{@{}l@{}}sadness, despair, giving up, \\ hopelessness, gloom\end{tabular}       & \includegraphics[scale=0.06]{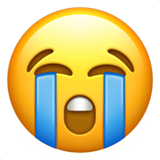}              \\ \hline
\textbf{disgust}     & a feeling of very strong dislike or disapproval.                                                                                                                       & \begin{tabular}[c]{@{}l@{}}loathing, dislike, hatred, \\ sicken, abomination\end{tabular}         & \includegraphics[scale=0.06]{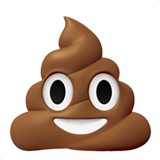}                \\ \hline
\textbf{excitement}  & \begin{tabular}[c]{@{}l@{}}a feeling of having great enthusiasm, strong belief, \\ intense enjoyment, or great eagerness.\end{tabular}                                 & \begin{tabular}[c]{@{}l@{}}enthusiasm, passion, \\ cheerfulness, heat\end{tabular}                & \includegraphics[scale=0.06]{Emoji/rocket_1f680.png}               \\ \hline
\textbf{optimism}    & \begin{tabular}[c]{@{}l@{}}a feeling of being hopeful about the future or \\ about the success of something in particular.\end{tabular}                                & \begin{tabular}[c]{@{}l@{}}hope, wish, desire,\\ want, positiveness\end{tabular}                  & \includegraphics[scale=0.06]{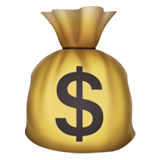}              \\ \hline
\textbf{panic}       & \begin{tabular}[c]{@{}l@{}}a very strong feeling of anxiety or fear, which makes you act \\ without thinking carefully.\end{tabular}                                   & \begin{tabular}[c]{@{}l@{}}horror, terror, fear, \\ dismay, terrify\end{tabular}                  & \includegraphics[scale=0.06]{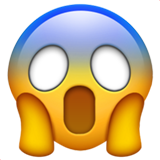}               \\ \hline
\textbf{surprise}    & \begin{tabular}[c]{@{}l@{}} a feeling caused by something that is unexpected or unusual. \\ (e.g. earning surprise)  \end{tabular}                                                                                                          & \begin{tabular}[c]{@{}l@{}}amazement, astonishment, \\ shock, revelation\end{tabular}             & \includegraphics[scale=0.06]{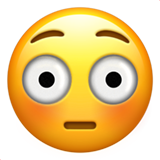}              \\ \hline
\textbf{ambiguous}   & \begin{tabular}[c]{@{}l@{}}unclassified emotions in the list or \\ when the target of emotion is confused.
\end{tabular} & \begin{tabular}[c]{@{}l@{}}(subject to annotator's \\ understanding of the text)\end{tabular}     & \includegraphics[scale=0.06]{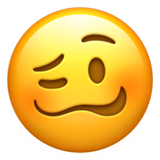}               \\ \hline
\end{tabular}
}
\caption{Emotion Definition given to the annotators.}
\label{tab:definition}
\end{table*}

\section{Dataset Details}
\subsection{Taxonomy of Emotions}
Inspired by the psychology of a market cycle \cite{hens2015behavioral}, we design our emotion taxonomy. In contrast to existing emotion classes that consolidate positive feelings into "\textit{joy}", it attempts to classify diversified mood proxies including hope, optimism, belief, and thrill. In order to maximise the impact of the dataset, we link our taxonomy with existing studies in psychology that serve Ekman’s 6-emotions \cite{ekman1992argument} (e.g. joy, surprise, disgust, fear, anger, and sadness) and Plutchik’s 8-emotions \cite{plutchik1980general}. Examining an emotion taxonomy in behavioral finance and psychology returns discrete emotions, but some are too similar. To address this, we make a subset of the data and ask two financial experts to annotate it with a predefined set of emotions. We notice that too similar labels such as {\itshape hope, wish, desire}, and {\itshape confidence} makes the annotation task more difficult thus, similar items are combined into a single representative emotion label (e.g. {\itshape optimism}).

\subsection{Annotation Guideline}
\label{sec:guide}
As shown in Table \ref{tab:definition}, emotion taxonomy, definition, synonyms, and related emoji are given to annotators. In addition, metadata including user sentiment, DateTime, and target company is provided to annotators in order to understand investors’ intentions. We clarify the taxonomy meaning through the seed and final round of annotation. For instance, {\itshape ambiguous} is also used for strong self-satisfaction or bragging about their success in a stock market downturn in order to tease opposing investors (e.g. with sarcasm). The initial taxonomy contains more labels such as {\itshape capitulation} and {\itshape denial}, but it is removed in the final round due to the small size of the data. If it is off-topic or unable to detect emotions, a {\itshape neutral} label is asked to annotate, and we discard it for the final dataset. 

\subsection{Grouping Emotions}
\label{sec:grouping}
For mapping Ekman’s 6 emotions, we use a similar approach to GoEmotions \cite{demszky2020goemotions}' mapping. Also, we use Plutchik's 8 emotions to group our taxonomy as follows:

To reduce noise, the ambiguous label is discarded. Mapping to Plutchik's emotions in StockEmotion.

\begin{itemize}
\setlength\itemsep{-0em}
\item \textbf{Grouping Ekman’s 6 emotions}: anger (maps to: anger), disgust (maps to: disgust), fear (maps to: anxiety, panic), joy (amusement, belief, excitement, optimism), sadness (maps to: depression) and surprise (all ambiguous emotions).
\item \textbf{Grouping Plutchik's 8 emotions}: anticipation (map to: optimism, confusion), anger (maps to: anger), disgust (maps to: disgust), fear (maps to: anxiety, panic), joy (amusement, excitement), sadness (maps to: depression), surprise (surprise, ambiguous), and trust (maps to: belief).
\end{itemize}

\section{Experimental Design}
\subsection{Experimental Setup}
For the classification task, we perform 5 independent experiment runs using different random seeds and present the average performance results across all runs. All experiment was running on 16GB T4 Nvidia GPUs. The final train/val/test split counts per label are presented in Table \ref{tab:split}.  

\begin{table}[t]
\centering
\small
\setlength{\tabcolsep}{3.5mm}{
\renewcommand{\arraystretch}{1.2}
\begin{tabular}{lr|rrr}
\hline
 & \textbf{total} & \textbf{train} & \textbf{val} & \textbf{test} \\ 
\hline
\textbf{bearish}     & 4526            & 3619                & 465           & 442             \\ 
\textbf{bullish}     & 5474            & 4381                & 535           & 558             \\ \hline
\textbf{ambiguous}   & 870            & 698                & 86          & 86             \\
\textbf{amusement}   & 817            & 651                & 83          & 83             \\ 
\textbf{anger}       & 386            & 309                & 38          & 39             \\ 
\textbf{anxiety}     & 1363              & 1099                & 132          & 132             \\ 
\textbf{belief}      & 907            & 728                & 89          & 90             \\ 
\textbf{confusion}   & 609            & 489                & 60          & 60             \\ 
\textbf{depression}  & 204            & 166                & 19          & 19             \\ 
\textbf{disgust}     & 1278            & 1036                & 121          & 121             \\ 
\textbf{excitement}  & 1380            & 1089                & 146          & 145             \\ 
\textbf{optimism}    & 1624            & 1299                & 163          & 162             \\ 
\textbf{panic}       & 302            & 240                & 31          & 31             \\ 
\textbf{surprise}    & 260            & 196                & 32          & 32             \\ 
\hline
\textbf{total}       & 10000            & 8000                & 1000          & 1000             \\ 
\hline
\end{tabular}
}
\caption{Split Details}
\label{tab:split}
\end{table}

\begin{table}[t]
\centering
\small
\setlength{\tabcolsep}{3mm}{
\renewcommand{\arraystretch}{1.2}
\begin{tabular}{ll}
\hline
\textbf{Model} & \textbf{hyperparameter search}  \\ 
\hline
\textbf{Logistic Reg.} & lr. [0.1, \textbf{0.01}, 1e-3]; ep. [5, \textbf{10}, 30]. \\ 
\textbf{NBSVM} & lr. [0.1, \textbf{0.01}, 1e-3]; ep. [\textbf{2}, 5, 10]. \\ \hline
\textbf{GRU} & lr. [0.1, 0.01, \textbf{1e-3}]; ep. [3, 6, \textbf{9}, 10]. \\
\textbf{Bi-GRU} & lr. [0.1, 0.01, \textbf{1e-3}]; ep. [3, \textbf{6}, 9, 10].\\ \hline
\textbf{DistilBERT} & lr. [1e-4, 1e-6, \textbf{3e-05}]; ep. [\textbf{3}, 5];\\ 
\textbf{BERT} &  lr. [1e-4, 1e-6, \textbf{3e-05}]; ep. [\textbf{3}, 5]; \\ 
\textbf{RoBERTa} & lr. [1e-4, \textbf{1e-6}, 3e-05]; ep. [\textbf{3}, 5]; \\
& (all) dropout [\textbf{0.2}, 0.5]. \\
\hline
\end{tabular}
}
\caption{Hyperparameter Configuration Details. (lr. = learning rate, ep. = epochs)}
\label{tab:hyperp}
\end{table}

\subsection{Baselines} 
\label{sec:baselines}
For classification baseline experiments, we compare machine learning methods, standard neural network and pretrained language models as follows:
\begin{itemize}
\setlength\itemsep{0em}
\item \textbf{Logistic Regression}: a probabilistic-based algorithm to predict the class with the highest probability. 
\item \textbf{Naïve Bayes SVM} \cite{wang2012baselines}: a variant of Support Vector Machines (SVM) using Naïve Bayes log-count ratios as feature values. 
\item \textbf{GRU} \cite{cho2014learning}: a gating mechanism in RNN, showing the empirical studies of better performance on less frequent dataset. 
\item \textbf{Bi-GRU}: a forward and a backward directional model consisting two GRUs.
\item \textbf{DistilBERT} \cite{sanh2019distilbert}: a light version of BERT for faster training. 
\item \textbf{BERT\(_{base}\)} \cite{DevlinCLT19}: a deeply pretrained language model using bidirectional encoder representations from Transformer. 
\item \textbf{RoBERTa\(_{base}\)} \cite{liu2019roberta}: a robustly optimized method for BERT by removing next sentence pretraining objective. 
\end{itemize}

\paragraph{Evaluation metrics}
We report per-class performances of an F1 score across all baselines as displayed in Table \ref{tab:eval}. Also, grouping emotions results using BERT on StockEmotions are shown in Table \ref{tab:eval_group6} and Table \ref{tab:eval_group8}.


\paragraph{Hyperparameter Search}
Each model's hyper-parameter settings are reported in Table \ref{tab:hyperp}. More variant results are obtained based on the change of learning rate. In bold, the best performing hyperparameter settings are shown.

\begin{table}[t]
\centering
\scriptsize
\small
\setlength{\tabcolsep}{2.5mm}{
\renewcommand{\arraystretch}{1.2}
\begin{tabular}{lccc}
\hline
\textbf{Ekman Emotion} & \textbf{Precision} & \textbf{Recall} & \textbf{F1} \\ \hline
anger                   & 0.54               & 0.64            & 0.59        \\ 
disgust                 & 0.40               & 0.38            & 0.39        \\ 
fear                    & 0.54               & 0.53            & 0.54        \\ 
joy                     & 0.74               & 0.78            & 0.76        \\ 
sadness                 & 0.17               & 0.21            & 0.19        \\ 
surprise                & 0.46               & 0.40            & 0.43        \\ \hline
\textbf{macro-avg.}     & \textbf{0.48}      & \textbf{0.49}   & \textbf{0.48} \\ \hline
\end{tabular}
}
\caption{Ekman's Mapping Results using BERT.}
\label{tab:eval_group6}
\end{table}

\begin{table}[t]
\centering
\scriptsize
\small
\setlength{\tabcolsep}{2.5mm}{
\renewcommand{\arraystretch}{1.2}
\begin{tabular}{lccc}
\hline
\textbf{Plutchik Emotion} & \textbf{Precision} & \textbf{Recall} & \textbf{F1}   \\ \hline
anticipation              & 0.53               & 0.54            & 0.53          \\ 
anger                     & 0.45               & 0.44            & 0.44          \\ 
disgust                   & 0.36               & 0.58            & 0.45          \\ 
fear                      & 0.47               & 0.55            & 0.51          \\ 
joy                       & 0.53               & 0.54            & 0.53          \\ 
sadness                   & 0.24               & 0.32            & 0.27          \\ 
surprise                  & 0.44               & 0.03            & 0.06          \\ 
trust                     & 0.41               & 0.41            & 0.41          \\ \hline
\textbf{macro-avg.}       & \textbf{0.43}      & \textbf{0.43}   & \textbf{0.40} \\ \hline
\end{tabular}
}
\caption{Plutchik's Mapping Results using BERT.}
\label{tab:eval_group8}
\end{table}

\subsection{Error Analysis}
We found that financial jargon and slang (e.g. ATH - All Time High; BTD - Buy The Deep; LMFAO - Laughing My Freaking Ass Off) generate confusion. For example, a \textit{positive} market movement is represented by \textit{long, call, green and bull}, while a \textit{negative} market movement is indicative by \textit{put, short, red and bear}. This type of jargon is frequently used across emotions, and it tends to be confused by the model.

\section{Limitations}
We are aware that the dataset contains potential biases including offensive languages, user base biases, and annotator biases because of the nature of data source. As the StockTwit platform is commonly used for promoting certain stocks, The quantity of content is skewed towards speculative or "meme" stocks, and the quality of content often contains toxic, offensive language, and sarcasm. In addition, the emotion of a message can often be confusing as investors express the opposite sentiment based on their investment position. For example, investors who hold Tesla stocks express their anger when the stock has fallen; however, their sentiment is annotated as bullish by implying their emotions of hope or belief. The collected data belongs to the era of COVID-19, so it would represent different results considering a much longer period (e.g. 10 years of stock market prediction). In order to present the impact of text and emotion combinations, a Temporal Attention LSTM architecture is presented and the comparison of other baseline remains for future work.

\end{document}